\documentclass[a4paper,conference]{IEEEtran}

\usepackage{times}
\usepackage{epsfig}
\usepackage{graphicx}
\usepackage{amsmath}
\usepackage{amssymb}
\usepackage{bm}

\usepackage{hhline}
\usepackage{multirow}

\newcommand{\figg}{{Figure~}}
\newcommand{\tabb}{{Table~}}

\begin{document}

\title{DAQN: Deep Auto-encoder and Q-Network}

\author{\IEEEauthorblockN{Daiki Kimura}
	\IEEEauthorblockA{IBM Research AI\\
		Email: daiki@jp.ibm.com}
}

\maketitle

\begin{abstract}
The deep reinforcement learning method usually requires a large number of training images and executing actions to obtain sufficient results. 
When it is extended a real-task in the real environment with an actual robot, the method will be required more training images due to complexities or noises of the input images, and executing a lot of actions on the real robot also becomes a serious problem. 
Therefore, we propose an extended deep reinforcement learning method that is applied a generative model to initialize the network for reducing the number of training trials.
In this paper, we used a deep q-network method as the deep reinforcement learning method and a deep auto-encoder as the generative model.
We conducted experiments on three different tasks: a cart-pole game, an atari game, and a real-game with an actual robot.
The proposed method trained efficiently on all tasks than the previous method, especially 2.5 times faster on a task with real environment images.
% when the input space and the policies are complex to be learned. 
%This method can reduce the number of training trials by pre-training the network.
\end{abstract}

% \fi
\IEEEpeerreviewmaketitle

% The maximum paper length is 8 pages excluding references and acknowledgements,
% 10 pages including references and acknowledgements
\section{Introduction}

The deep reinforcement learning algorithms, such as deep q-network method~\cite{Mnih2015, NIPS2014_5421}, deep deterministic policy gradients~\cite{ddpg}, and asynchronous advantage actor-critic~\cite{a3c}, are the suitable methods for implementing interactive robot~(agent). The method chooses an optimum action from a state input. These methods were often evaluated in a simulated environment. It is easy in the simulator to obtain input, to perform actions, and to get a reward. 
However, giving rewards to the real robot for each action needs human efforts. Also, performing training trials on the actual robot takes time and has risks of hurting the robot and environment. Thus, reducing the number of giving the reward and taking an action is necessary. 
Moreover, when we consider introducing into the real environment, state inputs will have a wider diversity. For example, if there is a state of ``in front of a human'', there are almost infinite variations of the visual representation of the human. Also, the input from an actual sensor contains complex background noises. Hence, a larger amount of training is essentially required than in the simulator environment.

On the other hand, a generative model has recently become popular to pre-train a neural network. An auto-encoder~\cite{Hinton504, Masci:2011:SCA:2029556.2029563} is one of the well-known generative model methods. Some studies, such as~\cite{Erhan:2010:WUP:1756006.1756025, 6639343}, report the auto-encoder reduced the number of training steps for the classification task. We, therefore, assume pre-training helps to reduce the number of reinforcement learning steps. Moreover, it only requires inputs during pre-training; class labels of data are unnecessary. Hence, the training data of the pre-training doesn't need rewards for reinforcement learning; that means we can obtain these data from a random policy agent, or the environment around the robot without any action. 
%Moreover, training a wide variety of inputs improves the robustness of the background noises.

\begin{figure}[t]
	\begin{center}
		\includegraphics[width=8.3cm]{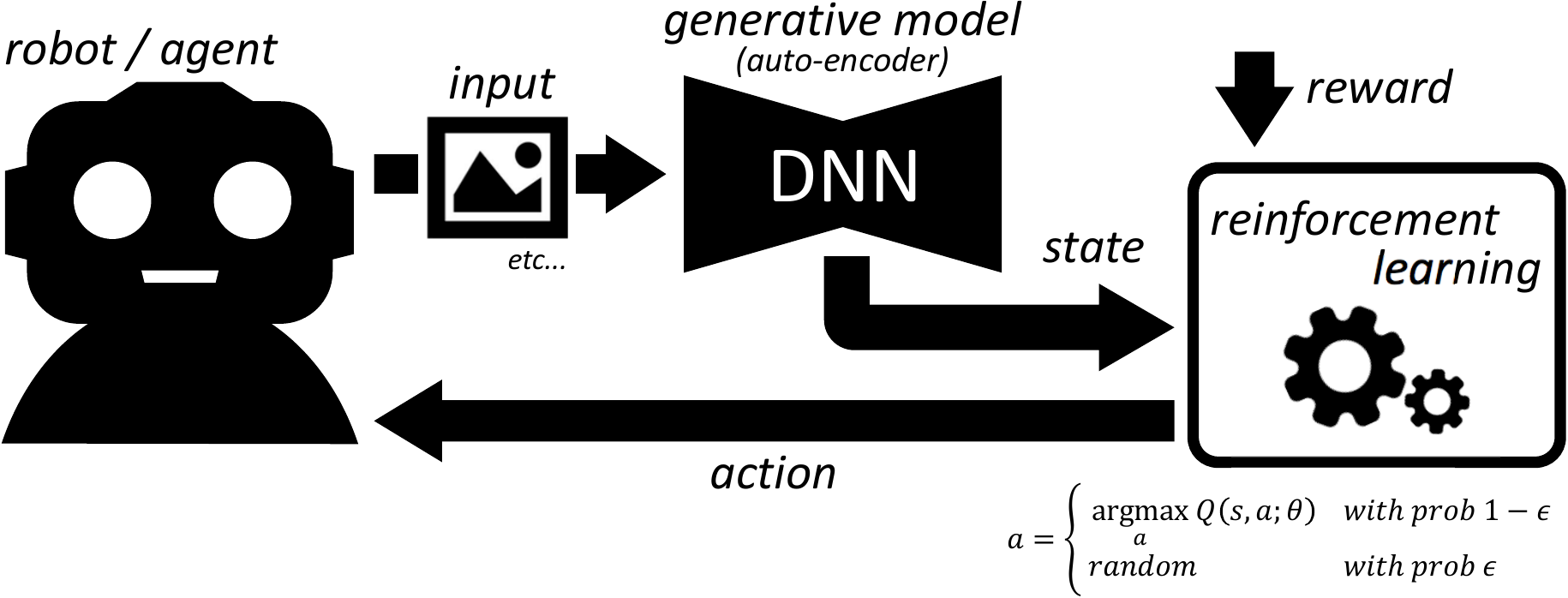}
	\end{center}
	\caption{Deep Auto-encoder and Q-Network}
	\label{fig:overview}
\end{figure}

Therefore, we proposed an extended deep reinforcement learning method that is applied a generative model to initialize the network for reducing the number of training trials. In this paper, we used a deep q-network method~\cite{Mnih2015, NIPS2014_5421} as deep reinforcement learning, and a deep auto-encoder~\cite{Masci:2011:SCA:2029556.2029563} as a generative model. We named this proposed method ``deep auto-encoder and q-network''~(daqn) in this paper. The overview is shown in \figg\ref{fig:overview}. 
This method first trains a network by the generative model with a random policy agent or inputs without actions. Then, the method trains policies by reinforcement learning which has the pre-trained network. The expected advantage is decreasing the number of training trials of the reinforcement learning phase. It is true that the proposed method additionally requires the training data for pre-training. However, the cost of obtaining this data is much lower than data for the reinforcement learning. This is because rewards and optimal actions are not required for the pre-training data. In this paper, we evaluate in three different environments, including a physical interaction with a real robot. Note that we focus on discrete actions in this paper for a clear discussion about contributions by a simple architecture. We assume the proposed method will be easily applied to continuous controls with such reinforcement learning methods~\cite{ddpg, a3c}. 
% Then we will discuss about the contributions of this method.

The contributions of this paper are to clarify the benefit of introducing the generative model into the deep reinforcement learning~(especially, the auto-encoder into the deep q-network), conditions for its effectiveness, and a requirement of pre-training data. 
The most similar work is~\cite{sandven2016visual}. They copied an auto-encoder network to deep q-network in atari environment. However, they concluded the pre-training results show lower performance. The main reason are it used only first layer to copy and some training parameters are not proper. We use all layers and change the parameters. And we conduct an experiment in the real environment that is more complex than the atari; thus, a benefit of pre-training will be expected more.

\section{Deep Auto-encoder and Q-Network}

The proposed method has following steps: (1) trains a network by the deep auto-encoder, (2) deletes the decoder-layers, and adds a fully-connected layer on the top of encoder-layers, and (3) trains policies by the deep q-network algorithm.

%\begin{figure}[t]
%\begin{center}
%\includegraphics[width=8.5cm]{fig/proposed_method.eps}
%\end{center}
%\caption{Deep Auto-encoder and Q-Network}
%\label{fig:proposed_method}
%\end{figure}

The method first trains inputs by a deep auto-encoder for pre-training the network. The auto-encoder has encoder and decoder components, which can be defined as transitions $\phi$ and $\psi$. Here, we define $\mathcal{X} = \mathbb{R}^n, \mathcal{Y} = \mathbb{R}^m$. Let $\phi^*$ and $\psi^*$ be trained by reconstructing its own inputs:
\begin{gather}
\phi :\mathcal{X} \rightarrow \mathcal{Y}, \:\: \psi :\mathcal{Y} \rightarrow \mathcal{X}, \:\: \bm{x} \in \mathcal{X} \\[5pt]
\phi^* ,\psi^* =\arg \min _{\phi ,\psi } \|\bm{x}-(\psi \circ \phi )\bm{x}\|^{2}.
\end{gather}
When the input is a simple vector, the proposed method uses a one-dimensional auto-encoder~\cite{Hinton504}; when the input is an image, it uses a convolutional auto-encoder~\cite{Masci:2011:SCA:2029556.2029563}. Importantly, the training data of this step will be obtained through a random policy of the agent, or the captured data from the environment without any action of the robot. 
%Note that the number and variety of training data will directly affect to the training effectiveness of a next reinforcement learning phase.

Next, the method removes decoder-layers of the auto-encoder network, and adds a fully-connected layer at the top of encoder-layers for discrete actions. Note that the weights of this added layer are initialized by random values.

Then, the method trains the policy by deep q-network~\cite{Mnih2015, NIPS2014_5421}, which is initialized by the pre-trained network parameters from previous steps. The deep q-network is based on Q-learning algorithm~\cite{watkins1992q}. The algorithm has an ``action-value function'' to calculate the quantity for a combination of state $S$ and action $A$; $Q: S\times A \to \mathbb {R}$. Then, the update function of $Q$ is,
\begin{multline}
Q(s_{t},a_{t}) \leftarrow \\
Q(s_{t},a_{t}) + \alpha \left( r_{t+1} + \gamma \max_{a}Q(s_{t+1},a) - Q(s_{t},a_{t}) \right),
\end{multline}
where the right-side $Q(s_{t},a_{t})$ is the current value, $\alpha$ is a learning ratio, $r_{t+1}$ is a reward, $\gamma$ is a discount factor, and $\max _{a}Q(s_{t+1},a)$ is a maximum estimated action-value for state $s_{t+1}$. 

Therefore, the loss function of the deep q-network is,

\begin{align}
L(\theta) &= \mathbb{E}_{(s_t,a_t,r_{t+1},s_{t+1})\sim\cal{D}} \bigl[ (y - Q(s_t,a_t; \theta) )^2 \bigl] \\[15pt]
y &= \left\{ \begin{array}{ll}
    r_{t+1} & \mbox{terminal}\\[7pt]
    r_{t+1} + \gamma \max_{a} Q(s_{t+1},a;\theta^{-}) & \mbox{non-terminal},
\end{array} \right.
\end{align}
where $\theta$ is the parameters of deep q-network, $\cal{D}$ is an experience replay memory~\cite{Lin:1992:RLR:168871}, $Q(s_t,a_t; \theta)$ will be calculated by the deep structure, and $\gamma$ is a discount factor. $\theta^{-}$ is the weights that are updated fixed duration; this technique was also used in the original deep q-network method~\cite{Mnih2015}.

%Then the method trains the policies by reinforcement learning from the acquired state value and a given reward. In this step, the method requires the sensor inputs and a reward resulting from an executed action.
%During this training the deep q-network, the input vector/image is converted to state values by the deep structure contains the pre-trained layers, and the method trains the policies by reinforcement learning with given reward, as shown in \figg\ref{fig:proposed_method}.

\section{Experiment}

%\renewcommand{\arraystretch}{1.2}
%\begin{table}[t]
%\begin{center}
%\begin{tabular}{ccc}
%% \begin{tabular}{c||c|c}
%\hline
% & Input & Game \\
% & complexity & complexity \\
%\hline
%\\[-11pt]
%% \hhline{=#=|=}
%\multirow{2}{*}{Cart-pole} & Simple & \multirow{2}{*}{Simple} \\
% & (4 values) & \\[8pt]
%% \hline
%\multirow{2}{*}{Atari} & Complex & \multirow{2}{*}{Complex} \\
% & (gray-image) & \\[8pt]
%% \hline
%\multirow{2}{*}{Real game} & Highly complex & \multirow{2}{*}{Simple} \\
% & (rgb-image with noise) & \\[8pt]
%\end{tabular}
%\end{center}
%\caption{Comparison Table of Games}
%\label{tab:comparison_games}
%\end{table}
%\renewcommand{\arraystretch}{1.0}

%In order to have a discussion about characteristic features of the proposed method, 
We conduct three different types of games in this study: a cart-pole game~\cite{6313077}, an Atari game which is implemented in the arcade learning environment~\cite{Bellemare:2013:ALE:2566972.2566979}, and an interactive game on an actual robot in the real environment. 
%\tabb\ref{tab:comparison_games} compares the complexities of games. 
The cart-pole is a simple game; hence, this is a base experiment. The Atari game, which is also used in the original deep q-network study or some related works, contains image inputs and a complex game rule. However, the images are taken from a simulated environment; specifically, images are from the game screen. The third environment is a real game. In this paper, we choose a ``rock-paper-scissors'' game, which is one of the various human hand-games with discrete actions. We adapt it to an interactive game between a real robot and a human. The input of this game is significantly complex than other experiments due to the real environment.
%We evaluate the proposed method in an actual environment The reason of preparing this game is to . The method that quickly trains a policy is required for extending the deep reinforcement learning to a real task, as mentioned in the introduction section. Proposing such a method is the targets of this work.
%Then we evaluate the proposed method with the original deep q-network.

We used the OpenAI Gym framework~\cite{1606.01540} for simulating the cart-pole game and the Atari game. Also, we used Keras-library~\cite{chollet2015keras} and Keras-RL~\cite{plappert2016kerasrl} for conducting the experiments.
%~\cite{tensorflow2015whitepaper}

\subsection{Cart-pole game}

\subsubsection{Environment}

The cart-pole game~\cite{6313077} is a game in which one pole is attached by a joint to a cart that moves along a friction-less track. The agent controls the cart to prevent the pole from falling over. It starts at the upright position, and it ends when the pole is more than 15 degrees from vertical.

\begin{figure}[t]
\begin{minipage}{0.48\hsize}
\begin{center}
	\includegraphics[width=4cm]{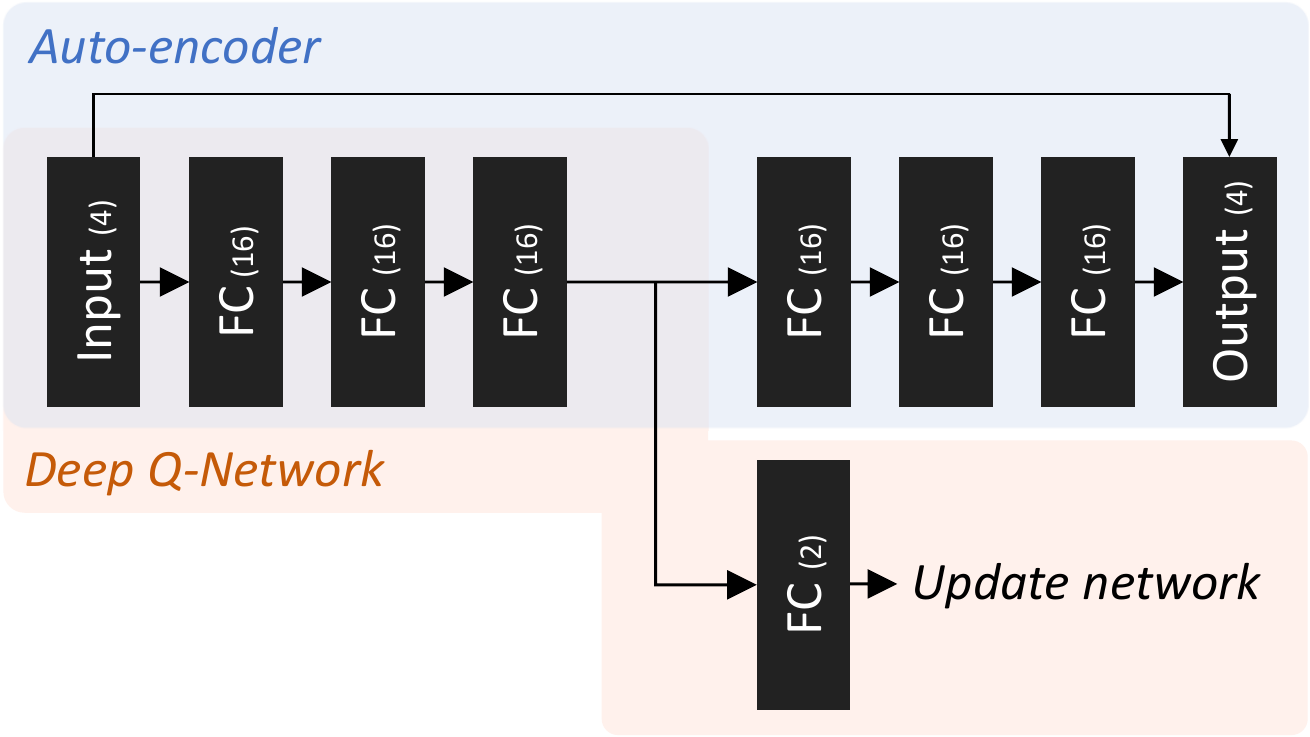}
\end{center}
\caption{Network for cart-pole game: blue part is auto-encoder component, orange part is deep q-network, and ``FC'' means a fully-connected layer.}
\label{fig:cartpole_proposed_method}
\end{minipage}
\begin{minipage}{0.5\hsize}
\begin{center}
\includegraphics[width=4.2cm]{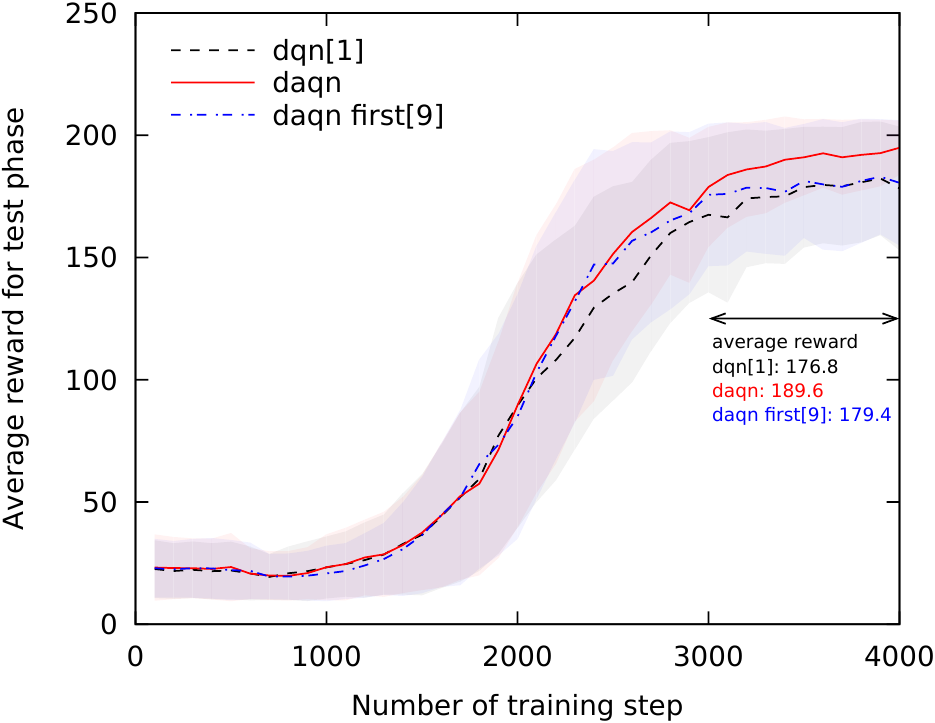}
\end{center}
\caption{Rewards for cart-pole game: the maximum reward is 300. Note that optimizer setting is only for auto-encoder~(ae); the optimizer of dqn is same as dqn method.}
\label{fig:cartpole_result}
\end{minipage}
\end{figure}

In this game, the agent obtains four-dimensional values from a sensor and chooses an optimal action from two discrete actions: moving right or left. To conduct an experiment, we first design a simple network that has three hidden layers and final layer for the deep q-network~(dqn). \figg\ref{fig:cartpole_proposed_method} shows the details of the whole network of the proposed method. The activation functions of all full-connected layers are a ReLU function~\cite{icml2010_NairH10}. The proposed method first pre-trains data from a random agent by auto-encoder algorithm; this is the blue part in \figg\ref{fig:cartpole_proposed_method}. Next, the method adds a full-connected layer at top of the encoder-layers. Then, it starts to train policies by using the dqn algorithm; this is the orange part in \figg\ref{fig:cartpole_proposed_method}. We compare the training efficiency with the original dqn method that doesn't have pre-training. Note that ``training step'' in the evaluation is for dqn component, excludes the number of auto-encoder training. Because the pre-training data is acquired from a random policy; thus it doesn't require much cost.
%Note that the difference between the original deep q-network and proposed method is training from scratch or training from the pre-trained network.

The pre-training data is inputs of 10 thousand~(10k) steps from a random policy. The loss of auto-encoder is a mean square error function, the optimizer is an adaptive moment estimation~(adam)~\cite{ICLR:journals/corr/KingmaB14}, and we train $25$ epochs. Also, we try the \cite{sandven2016visual} method that uses only first layer of the pre-trained network to copy to dqn. However, we use new parameters that are same as our method; some are different from \cite{sandven2016visual}. The optimizer of dqn component is adam, and exploration policy is Boltzmann approach. Note that all parameters and settings of the dqn component in daqn are same as those in the original dqn method.
%We changed an optimizer for the pre-training to evaluate the stability. We tried adaptive moment estimation (adam)~\cite{ICLR:journals/corr/KingmaB14} and some fixed learning rates of stochastic gradient descent (sgd).

\subsubsection{Results}

\figg\ref{fig:cartpole_result} shows the reward curves for test phase with different numbers of training steps. Results were computed from running 100 episodes, 10 different training-trials; then the line is average of these results. The average rewards after 3k training steps were 176.8, {\bf 189.6}, and 179.4, for the dqn, the proposed method~(daqn), and \cite{sandven2016visual} which uses only first layer, respectively. These results show pre-training by auto-encoder was efficient to improve the reward.
%However, the effectiveness depends on the type of optimizer for the auto-encoder component, which sometimes has similar results as those from the method has deep q-network component only.

\subsection{Atari game}

\subsubsection{Environment}

% \begin{figure}[t]
% \begin{center}
% \includegraphics[width=7cm]{fig/atari_proposed_method_1.eps}
% \end{center}
% \caption{Network (a) of proposed method for Atari game}
% \label{fig:atari_proposed_method_1}
% \end{figure}

\begin{figure}[t]
\begin{minipage}{0.48\hsize}
\begin{center}
\includegraphics[width=4cm]{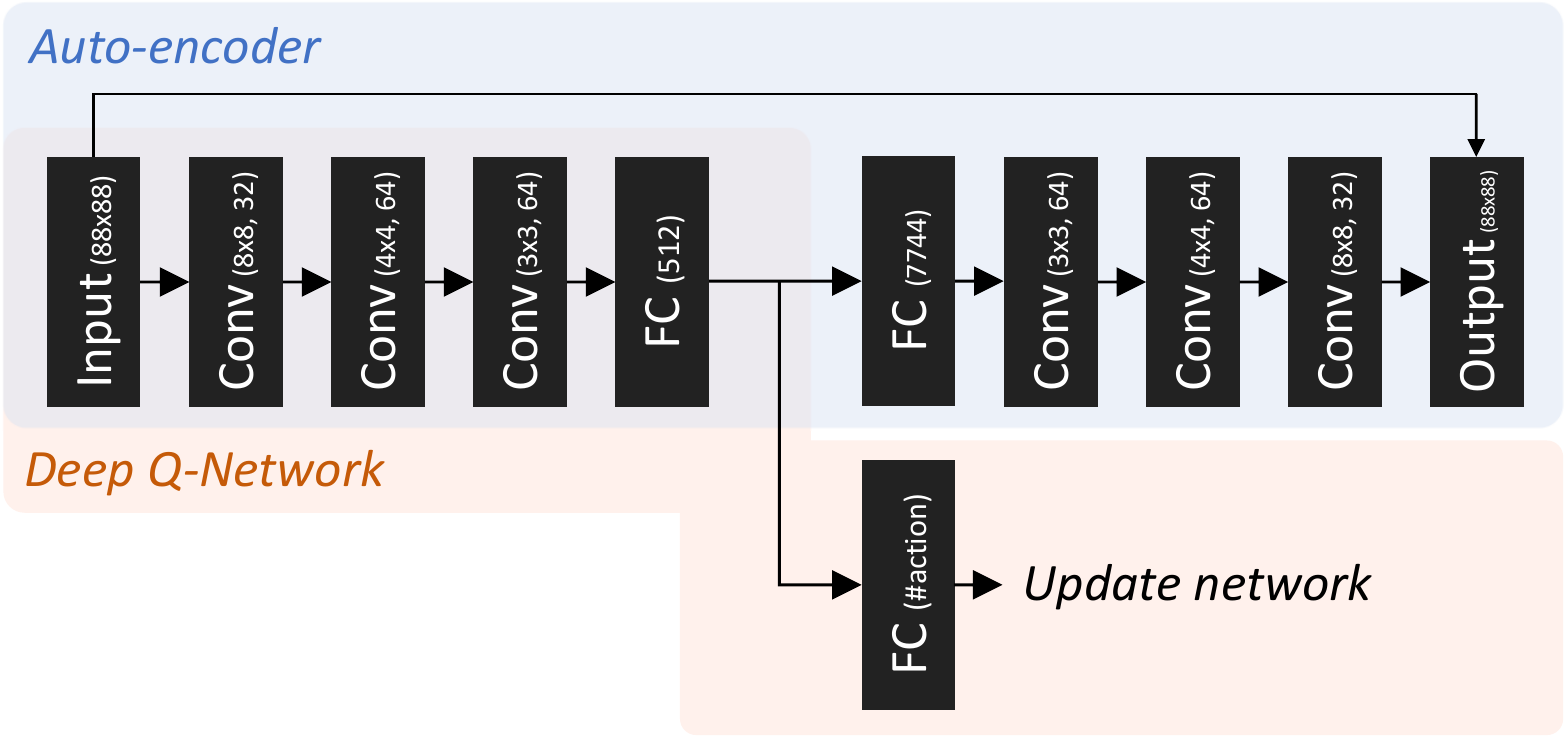}
\end{center}
\caption{Proposed network for Atari game: ``Conv'' means a convolutional layer.}
\label{fig:atari_proposed_method_2}
\end{minipage}
%\end{figure}
%\begin{figure}[t]
\begin{minipage}{0.5\hsize}
\begin{center}
\includegraphics[width=4cm]{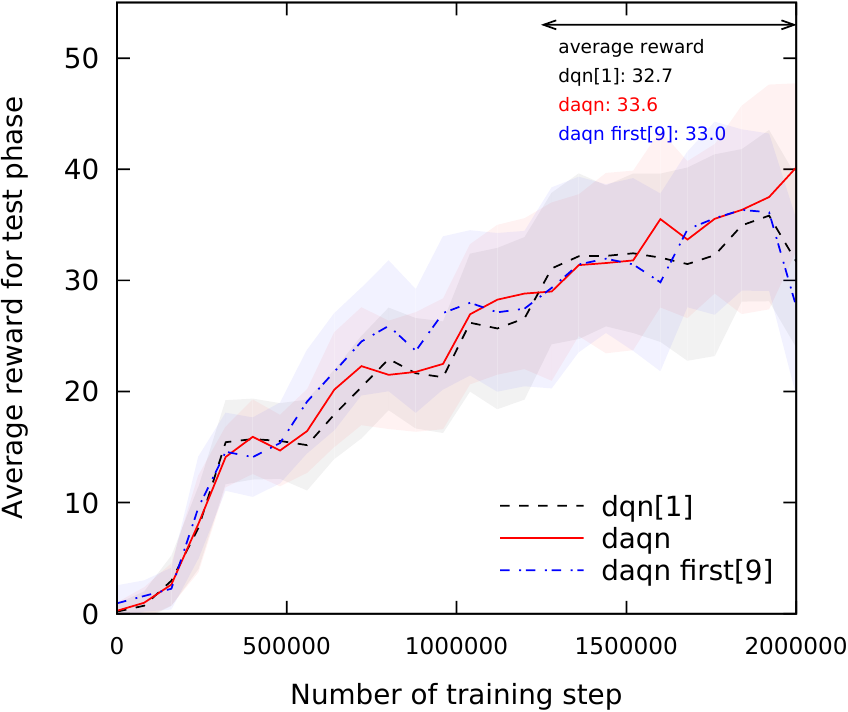}
\end{center}
\caption{Average rewards for Breakout.}
\label{fig:atari_result}
\end{minipage}
\end{figure}

% \begin{figure}[t]
% \begin{center}
% \includegraphics[width=8cm]{fig/atari_proposed_method_2.eps}
% \end{center}
% \caption{Network of proposed method for Atari game}
% \label{fig:atari_proposed_method_2}
% \end{figure}

%, and it has recently become a common evaluating environment for the deep reinforcement learning.
%Atari game contains many arcade games. The agent will choose optimal actions based on the acquired game-screen images. 
In the previous study~\cite{Mnih2015, NIPS2014_5421}, they proposed a trainable deep convolutional neural network. We use the same network for this proposed method. However, we change the initial image size from $84 \times 84$ to $88 \times 88$, due to adding an auto-encoder component. All other parameters and settings were the same as those in the previous study. 

We choose ``Breakout'' for evaluation; because it's one of the games in which the dqn trained successfully~\cite{Mnih2015}. \figg\ref{fig:atari_proposed_method_2} shows the proposed network. The pre-training data is 100k step images which are captured by a random policy agent. We use a fixed learning rate~(0.01) with stochastic gradient descent for auto-encoder component, and train $25$ epochs. Also, we try the~\cite{sandven2016visual} method which uses only first layer to copy; parameters are same as the proposed method, thus different from \cite{sandven2016visual}.
%However, the optimizer, exploration policy, and other settings of deep q-network component are same as original dqn method.
%We tried adam-optimizer and several fixed learning rates (lr) of stochastic gradient descent~(sgd).
%However, we did not change the settings of the deep q-network component.

\subsubsection{Results}

% \begin{figure}[t]
% \begin{center}
% \includegraphics[width=8.3cm]{fig/table_daqn_atari_tst_breakout0_m20170216224219ONE_t1000.eps}
% \end{center}
% \caption{Average rewards for Breakout in Atari game}
% \label{fig:atari_result}
% \end{figure}

% \begin{figure}[t]
% \begin{center}
% \includegraphics[width=8.3cm]{fig/daqn_atari.pdf}
% \end{center}
% \caption{Average rewards for Breakout in Atari game}
% \label{fig:atari_result}
% \end{figure}

\figg\ref{fig:atari_result} shows the reward curves with different numbers of training steps. The statistics were computed from running 20 episodes. The average rewards after training 1.25~million steps were 32.7, {\bf 33.6}, and 33.0, for dqn, daqn, and \cite{sandven2016visual}, respectively. These results show the proposed method has the best result; however, it is similar to the dqn result.

% 1900000:-1
% dqn_atari_trntst_vBreakoutDeterministic-v4_s2000000_i10000_m20170620: 34.5866666667
% daqn_atari_trntst_vBreakoutDeterministic-v4_lae_atari_trn2_vBreakoutDeterministic-v4_dn100000_ncacacafdadarucaucaucac_b128_e25_r1_oadam_m20170615000000mse_s2000000_i10000_m20170620: 33.74
% daqn_atari_trntst_vBreakoutDeterministic-v4_lae_atari_trn2_vBreakoutDeterministic-v4_dn100000_ncacacafdadarucaucaucac_b128_e25_r1_osgd0.01_m20170615000000mse_s2000000_i10000_m20170620: 34.7966666667

\subsection{Real-world game}

\subsubsection{Game setting}

In this section, we conduct an experiment to evaluate the proposed method in the real environment. However, there is no suitable previous game-task with such environment. Hence, we choose a rock-paper-scissors game from some interactive human-games based on discrete actions. The optimal action of reinforcement learning is to beat a human-opponent by getting an opponent's hand image. The possible actions are to make a posture which represents `rock', `paper', and `scissors'.
%During this training Therefore, the robot must train a policy to win from a noisy image that are taken in the real environment.
%has highly complex input and simple game rules. Because we assumed that the effectiveness of the auto-encoder could be increased by using a large variety of input images. Furthermore, the interactions with a real robot will be advancing the applications of reinforcement learning technology(?later).
%It means the agent just needs to learn the best selection for each image input. However compared to other experiments, it uses the actual camera images with complex background. Therefore the complexity of image is higher than other experiments. This difficulty is suit for advantages of auto-encoder.

\subsubsection{Pre-experiment}

\begin{figure*}[t]
\begin{center}
\includegraphics[width=17cm]{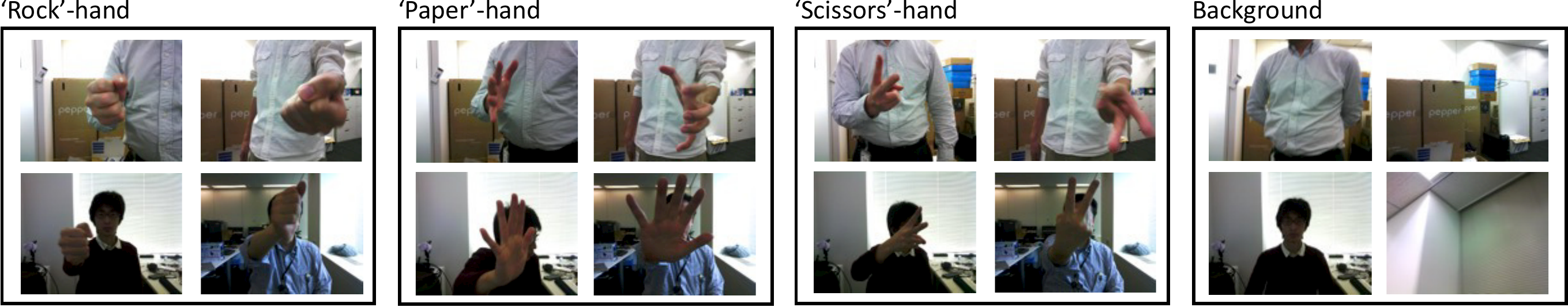}
\end{center}
\caption{Sample images for rock-paper-scissors game}
\label{fig:janken_sample}
\end{figure*}

\begin{table}[t]
\begin{center}
\begin{tabular}{ccccc}
\hline
 i & ii & iii & iv & v \\
\hline
%\multicolumn{5}{c}{input (resize to 120 x 120)} \\
input & input & input & input & input \\
% \hline
conv & conv & conv & conv & conv \\
% \hline
max & max & max & max & max \\
% \hline
conv & conv & conv & conv & conv \\
% \hline
max & conv & conv & conv & conv \\
% \hline
fc & fc & max & max & max \\
% \hline
softmax &softmax & fc & conv & dropout \\
% \hline
& & softmax & max & conv \\
% \hline
& & & fc & max \\
% \hline
& & & softmax & dropout \\
% \hline
& & & & fc \\
% \hline
& & & & fc \\
% \hline
& & & & softmax \\
\hline
\\[-8pt]
95.7\% & 97.2\% & {\bf 97.8\%} & 97.6\% & 97.1\% \\
\end{tabular}
\end{center}
\caption{Networks and accuracy for classification task.}
\label{tab:janken_networks}
\end{table}

\begin{figure}[t]
\begin{center}
\includegraphics[width=8cm]{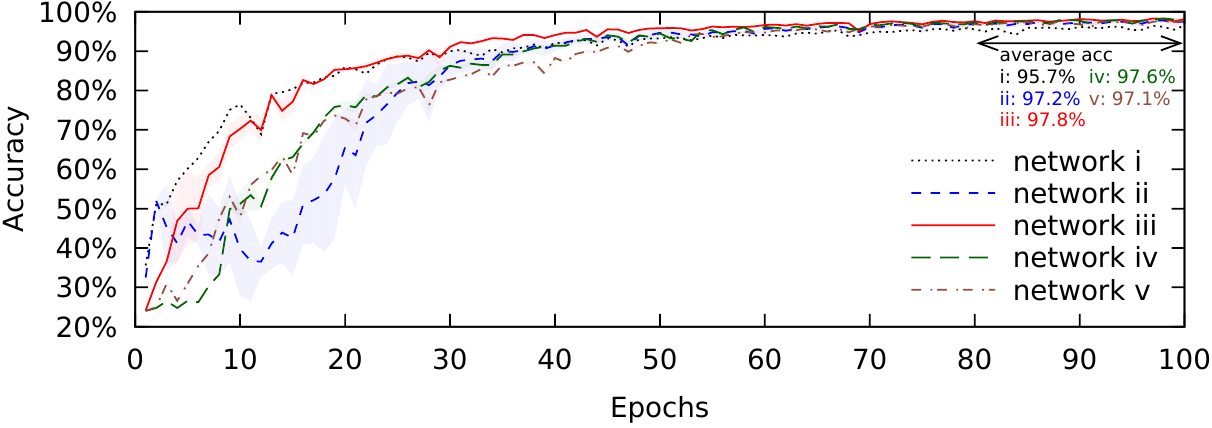}
\end{center}
\caption{Accuracy curve of classification}
\label{fig:janken_cnn_result}
\end{figure}

First, we must find a deep network that can do classification for these hand images. We take totally 50k images of four types of images: `rock'-hand, `paper'-hand, `scissors'-hand, and background. During this shooting, we change a person, a hand~(right or left hand), a height of the hand, wearing clothes, background environment, and a light condition. \figg\ref{fig:janken_sample} shows examples of taken images. Furthermore, these hand gesture images are not only the `completed' figure but also the figure during the changing hand, such as a right-up image of the `paper' in \figg\ref{fig:janken_sample}. We prepare some networks to classify into four classes. The upper part of \tabb\ref{tab:janken_networks} describes the detail of networks. Images were initially cropped to a $120 \times 120$ color image. Each convolutional and fully-connected layer has the ReLU activation function. Training~(90\%) and test~(10\%) images were separated.
%The reason to prepare no-hand images is we assume taking images for the pre-training of the deep q-network will contain such kind of images. 

\figg\ref{fig:janken_cnn_result} shows accuracy curve from five different trials, and the final row of \tabb\ref{tab:janken_networks} shows the average accuracy after 80 epochs. According to these results, network~iii gave the highest accuracy, we will use this network structure.

\subsubsection{Environment}

\begin{figure}[t]
\begin{center}
\includegraphics[width=8cm]{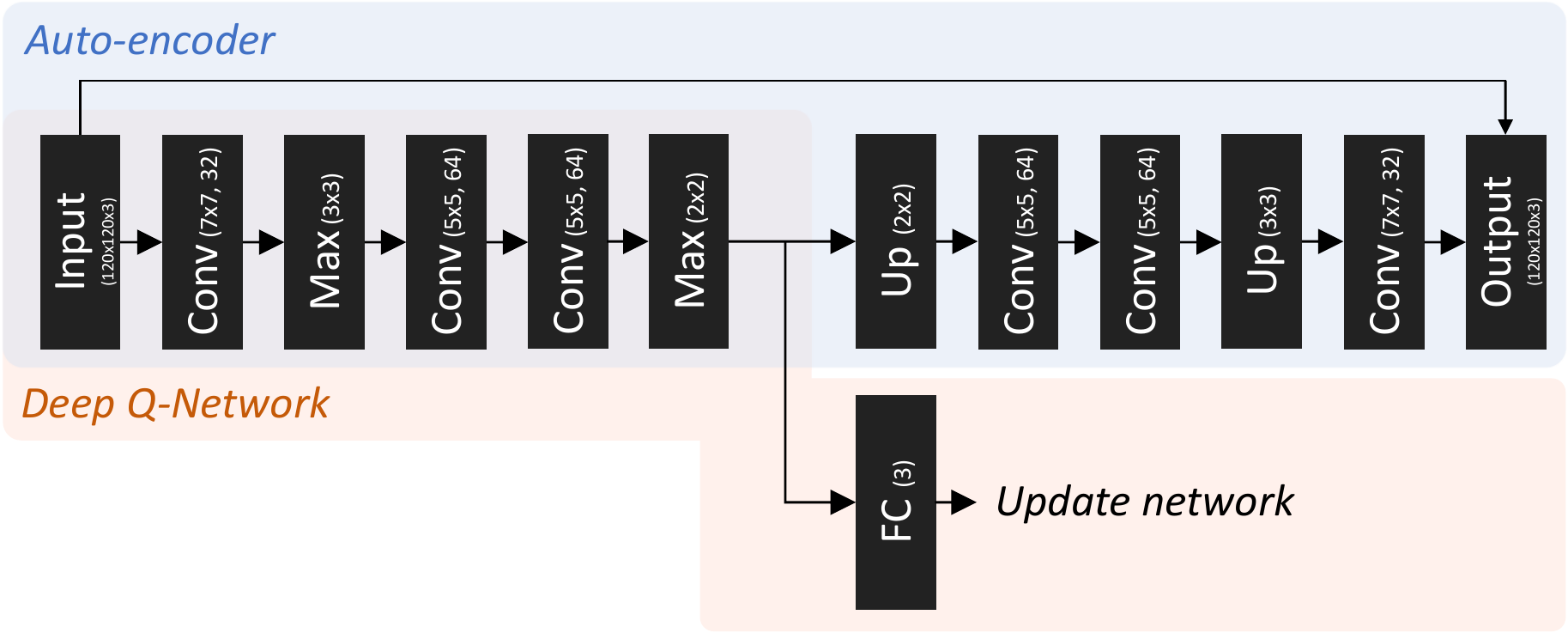}
\end{center}
\caption{Proposed network for real game: ``up'' is up-sampling.}
\label{fig:janken_proposed_method}
\end{figure}

%\begin{figure}[t]
%\begin{center}
%\includegraphics[width=8.3cm]{fig/output_autoencoder.eps}
%\end{center}
%\caption{Sample result of auto-encoder}
%\label{fig:output_autoencoder}
%\end{figure}

\begin{figure}[t]
\begin{center}
\includegraphics[width=8.5cm]{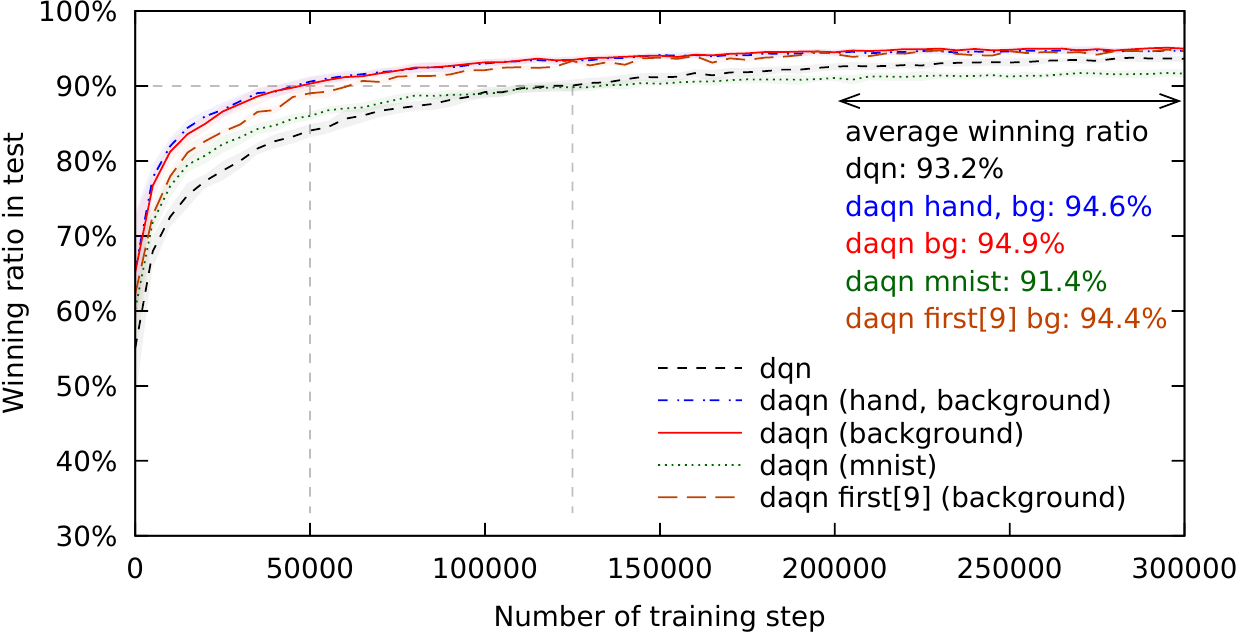}
\end{center}
\caption{Winning ratio curve for the rock-paper-scissors game}
\label{fig:janken_result}
\end{figure}

\figg\ref{fig:janken_proposed_method} shows the proposed method with network~iii. In order to clear a requirement of pre-training data, we prepare the following three types of pre-training data: all images that are taken in pre-experiment~(three hand-types and background), only background images, and images of handwritten digits~(mnist)~\cite{lecun-98-mnist}. If the proposed method can train from only the background images, the cost to take the image is lower than hand images. Because we just require taking around the robot's environment; the human hand is not necessary. And, of course, if we can train from mnist, it is the lowest cost to prepare. We train 30 epochs for the auto-encoder component with an adam optimizer, random image shifting, rotating, and flipping.

%\figg\ref{fig:output_autoencoder} shows sample of reconstructed images by the auto-encoder for the dataset of three hand-types and background. 
After the auto-encoder training, it trains policies by dqn component. We use only hand-types images from taken images in the previous experiment; background images were excluded. The reward is $+1$, $0$, or $-1$, when the agent wins, draws, or loses, respectively. The settings of dqn method and dqn component in daqn are as follows: learning ratio is $0.001$ with sgd, size of replay memory is $1000$, batch size is $32$, exploration policy is $\epsilon$-greedy, and loss is a mean square error function. And we compare with the \cite{sandven2016visual} method; training parameters are same as daqn.

\subsubsection{Results}

% \begin{figure}[t]
% \begin{center}
% \includegraphics[width=8.5cm]{fig/table_qlearn_daqn_long_bc.eps}
% \end{center}
% \caption{Winning ratio curve for the real-world game}
% \label{fig:janken_result}
% \end{figure}

\figg\ref{fig:janken_result} shows a graph for the winning ratio curve of test images with different numbers of training steps. Each statistic was computed from 4k test-trials, and 5 different training-trials. The average winning ratios after 200k training are 93.2\%, 94.6\%, {\bf 94.9\%}, 91.4\%, and 94.4\% for dqn, daqn which is pre-trained images of hand-types and background, daqn pre-trained background images, daqn pre-trained mnist images, and \cite{sandven2016visual} method pre-trained background images, respectively. According to these results, the proposed method can train faster than the dqn method; for example, it is around 2.5 times faster to reach a 90\% winning ratio. Also, the background images are enough to pre-train the network. Although mnist images initially help the improvement, the final winning ratio is lower than the dqn method. It means this pre-training data misled feature extraction in the reinforcement learning. Therefore, the requirement of pre-training data is these must have come from a domain related environment; however, backgrounds of the environment are accepted. The processing time for 300k training that includes evaluating was around 16.7 hours with Nvidia K40, and one test step took around 5.01 milli-seconds with Nvidia GTX Titan.

Moreover, we implement this game to an actual robot with the daqn pre-trained background. We attach a video that shows each robot reaction with a different number of training trials. In the video, the robot wins only one time at first. It can win gradually, it finally beats all types of opponent hands.

\section{Discussions}

\renewcommand{\arraystretch}{1.4}
\begin{table}[b]
\begin{center}
\begin{tabular}{cccc}
\hline
 & Input & Game & \multirow{2}{*}{DAQN} \\ [-3pt]
 & complexity & complexity & \\
\hline
\\[-11pt]
Cart-pole & Simple & Simple & $>$ DQN \\[6pt]
Atari & Complex & Complex & $\simeq$ DQN \\[6pt]
Real game & Highly complex & Simple & $\gg$ DQN \\[6pt]
\end{tabular}
\end{center}
\caption{Conclusion of these experiments}
\label{tab:discussion}
\end{table}
\renewcommand{\arraystretch}{1.0}

We applied the proposed method to different games: a simple game, a complex game in a simulator, and a real game. \tabb\ref{tab:discussion} shows the conclusion of these experiments. The proposed method has a pre-training component for the input; therefore, the complexity of the input directly affects the advantage of the proposed method. Hence, the most effective case is the game in the real environment. On the other hand, although the proposed method has a better result, the contribution is not so high in the atari game. We guess the main reason is a complexity of game. The pre-training data is based on the random policy; hence, images of the game did not change very much due to the difficulty of the game. Therefore, the pre-trained network parameters of daqn are easy to be acquired during initial steps of the dpn. Hence, the complexity of the game is also important for this pre-training.

% We concludes the proposed method performs efficient with complex sensory inputs and simple game rules.
%Moreover, the training the similar inputs during the pre-training easily leads to reach the local optimal solutions. For example, the proposed method with adam-optimizer of the Breakout game took a slow start of reward improvement, as shown in \figg\ref{fig:atari_result}.
%On the other hand, the proposed method also performed well for cart-pole game although the complexity of inputs is simple. Because the agent could get the various inputs even if it took random actions.
%It is preferable to obtain the input from a dynamic environment, which means the proposed method performs efficiently with a noisy background image or sound. Hence, the proposed method will be effective for actual robots in human environments.

Although it is less, the number of training trials of the proposed method is still large. It is still difficult to implement a system that will try ``live training'' on a physical robot. The reason for this long training is the difficulty to train inputs from the real environment. If we can use an improved deep reinforcement learning method, we hope it will converge faster; however, the processing time for test phase is also important for a real robot. 
%The reason of this long training is the difficulty to train policies. We believe a method that can roughly pre-train policies will be proposed. Moreover, the original deep q-network method improved according to recent studies~\cite{DBLP:journals/corr/HasseltGS15, ddpg, DBLP:journals/corr/WangFL15}. We expect improved deep q-network methods will help decrease this number of training trials.

\section{Related work}

The deep q-network method~\cite{Mnih2015, NIPS2014_5421} is a method that successfully applies a deep convolutional neural network~\cite{lecun2015deep} to reinforcement learning~\cite{sutton1998reinforcement} for training a policy of computer games. The convolutional neural network was originally inspired from neocognitron~\cite{Fukushima1980}, then it has become a well-known approach for extracting high-level features from a raw image. This has especially led to significant breakthroughs in image processing studies~\cite{NIPS2012_4824, 7298594, He_2016_CVPR}.

Reinforcement learning~\cite{sutton1998reinforcement} is an approach for training action policies with maximizing future cumulative rewards. There are several algorithms, such as Q-learning~\cite{watkins1992q} and TD-gammon~\cite{Tesauro:1995:TDL:203330.203343}. The TD-gammon applied a multi-layer perceptron for calculating approximated values, this achieved a human level to play a backgammon game.

The deep q-network~(dqn)~\cite{Mnih2015, NIPS2014_5421} has a deep structure network that calculates the q-value of \cite{watkins1992q} from a raw input. The neural fitted q-learning~(nfq)~\cite{nfq2005} also had a similar mechanism; however, the nfq used a batch update that has a computational cost per iteration. Also, the dqn used stochastic gradient updates that have a low constant cost per iteration~\cite{NIPS2014_5421}. Moreover, dqn has the latest convolutional neural network.
%And the deep q-network method uses the experience-replay technique~\cite{Lin:1992:RLR:168871}.
%The contribution of the deep q-network method is that it successfully trains control policies by using the convolutional neural network and reinforcement learning.

The deep auto-encoder was proposed for the dimensionality reduction~\cite{Hinton504}. However, it has also become more widely used for learning a generative model of the data~\cite{Erhan:2010:WUP:1756006.1756025, 6639343}. Also, the original deep auto-encoder~\cite{Hinton504} has a one-dimensional structure. However, it is difficult to preserve the spatial locality information of images with this structure. Therefore, a convolutional auto-encoder was proposed~\cite{Masci:2011:SCA:2029556.2029563}, and some studies applied to various problems~\cite{NIPS2016Mao, oyedotun2016pattern, DBLP:journals/corr/TurchenkoL15}. 
%We also use this latest method, when the input is an image.

A study to combine the spatial auto-encoder and reinforcement learning for real robot was proposed~\cite{ftddla-dsae-15}, they used the spatial auto-encoder for understanding the camera input from the robot. However, they used the spatial auto-encoder for describing the environment, such as the positions of objects, this is not for pre-training the network. Another reinforcement learning study with the auto-encoder was also proposed~\cite{DBLP:journals/corr/StadieLA15}, they introduced the additional reward, which likes ``curiosity''~\cite{pathakICMl17curiosity}, by the auto-encoder. However, they used auto-encoder as a method for reducing the dimension of state input, which is same application example as the original deep auto-encoder~\cite{Hinton504}.

The similar study with the proposed method is the deep auto-encoder neural networks in reinforcement learning~\cite{5596468}. This method first trains features by the auto-encoder, then trains policies by the batch-mode reinforcement learning algorithm. The architecture of this method is similar to ours; however, the motivation is different. They used the auto-encoder for reducing the dimension of input for the reinforcement learning. We used the auto-encoder to reduce the number of training trials for the reinforcement learning phase. Hence, we discussed the training efficiency in this paper. Furthermore, we used the latest convolutional auto-encoder and the latest deep reinforcement learning method.

The pre-training method with neural fitted q-learning was proposed~\cite{Abtahi:2011:DBN:2908756.2908757}. They tried to pre-train from the completely random policy, and ``hint-to-goal heuristic'' policy. They reported pre-training without ``hint'' seems to do nothing. We believe the reason is the input complexities; they only tried the Mountain Car and Puddle World which are simple. In this paper, we discussed on the more complex environment.

The most similar study is~\cite{sandven2016visual}. They copied pre-trained network parameters by auto-encoder to deep q-network, and they evaluated on atari games. However, they concluded ``the results generally show lower performance for cases with pre-training''. We think this has three reasons: type of pre-trained data, structure of copied network, and hyper-parameters for training. First, they mainly focused on transferring trained network by multiple games or imagenet; they called ``SSAE'' and ``INAE''. These results are expected to become bad; this is same as the mnist result in this paper. Second, they only copied the pre-trained parameters of first layer; this is not so sufficient to help the dqn. They didn't try to copy ``all'' network of auto-encoder to dqn in ``GSAE'' case. Third, when we tuned the some hyper-parameters for training, the pre-training helped slightly even if we use their method on atari game. And the big difference with their paper is they only discussed atari; however, we conducted an experiment of the real game that is expected to have a good advantage of pre-training. 

A robot to beat a human at rock-paper-scissors was proposed~\cite{7743496}, they constructed a new active sensing system that can actively track and recognize the human hand by using a high-speed vision system. However, they didn't use the reinforcement learning, thus it could not train the optimal action. And they used a heuristic algorithm to understand the human hand, it is not robust in the noisy environment. 

% Moreover, they only tried applying a simple navigation task. We discuss experiments with more complex and realistic tasks.
%in which the auto-encoder was applied for performing reinforcement learning from the raw input(? later).

\section{Conclusion}

We proposed an extended deep reinforcement learning that is applied one of the generative models to reduce the number of training steps. We evaluated it by three different games: a basic cart-pole game, a well-known atari game, and a ``rock-paper-scissors'' game in the real environment. Our method could train efficiently in all conditions; it especially works well when the input image has high complexity. For example, it trained 2.5~times faster than the original deep q-network method. Also, it could train from the background images which can be easily taken. We expect introducing the generative model must be required the deep reinforcement learning on actual robots in the real environment.

{\small
\bibliographystyle{IEEEtran}
\bibliography{ms}
}

\end{document}